\documentclass[conference]{IEEEtran}
\IEEEoverridecommandlockouts
\usepackage{cite}
\usepackage{amsmath,amssymb,amsfonts}
\usepackage{algorithmic}
\usepackage[ruled]{algorithm2e}
\usepackage{graphicx}
\usepackage{textcomp}
\usepackage{multirow}
\usepackage{xcolor}
\usepackage[hidelinks]{hyperref}

\usepackage{subcaption}
\usepackage{makecell}

\usepackage{optidef}
\def\BibTeX{{\rm B\kern-.05em{\sc i\kern-.025em b}\kern-.08em
    T\kern-.1667em\lower.7ex\hbox{E}\kern-.125emX}}
\begin{document}

\title{Effects of Different Optimization Formulations in Evolutionary Reinforcement Learning on Diverse Behavior Generation\\
}

\author{\IEEEauthorblockN{Victor Villin}
\IEEEauthorblockA{\textit{Dept. of Computer Science and} \\
\textit{Intelligent Systems} \\
\textit{Osaka Prefecture University}\\
Sakai, Osaka, Japan \\
villinvict@ci.cs.osakafu-u.ac.jp}
\and
\IEEEauthorblockN{Naoki Masuyama}
\IEEEauthorblockA{\textit{Dept. of Computer Science and} \\
\textit{Intelligent Systems} \\
\textit{Osaka Prefecture University}\\
Sakai, Osaka, Japan \\
masuyama@cs.osakafu-u.ac.jp}
\and
\IEEEauthorblockN{Yusuke Nojima}
\IEEEauthorblockA{\textit{Dept. of Computer Science and} \\
\textit{Intelligent Systems} \\
\textit{Osaka Prefecture University}\\
Sakai, Osaka, Japan \\
nojima@cs.osakafu-u.ac.jp}
}

\maketitle

\begin{abstract}
Generating various strategies for a given task is challenging. However, it has already proven to bring many assets to the main learning process, such as improved behavior exploration. With the growth in the interest of heterogeneity in solution in evolutionary computation and reinforcement learning, many promising approaches have emerged. To better understand how one guides multiple policies toward distinct strategies and benefit from diversity, we need to analyze further the influence of the reward signal modulation and other evolutionary mechanisms on the obtained behaviors. To that effect, this paper considers an existing evolutionary reinforcement learning framework which exploits multi-objective optimization as a way to obtain policies that succeed at behavior-related tasks as well as completing the main goal. Experiments on the Atari games stress that optimization formulations which do not consider objectives equally fail at generating diversity and even output agents that are worse at solving the problem at hand, regardless of the obtained behaviors.
\end{abstract}


\section{Introduction}
As we saw development in the Reinforcement Learning (RL) and Evolutionary Computation (EC) fields \cite{mnih_asynchronous_2016, jaderberg_population_2017, such_deep_2018}, we also noticed the emergence of approaches that would no longer solely pursue performance, but try to also consider diversity in solution as another goal.

This progressive turn is broadly motivated by the fact that problems currently tackled by RL tend to be hard to explore due to their high deceptiveness. Indeed, the aim of diversity is to counteract premature convergence issues and to improve exploration of the behavior space by allowing discovery of new policies. Moreover, achieving diversity is appealing as it holds many other benefits. It has short-term advantages with the learning process, allowing higher quality training with improved stability and final results. Long-term advantages are the acquisition of diverse and optimal solutions which can be used as alternative options to solve a task, or as different ways to adapt to dynamic multi-agent environments.

The recurrent problem of RL approaches is caused by their converging nature (i.e., future expected reward maximization). They tend to be subject to premature convergence and low policy exploration. Various studies to enhance exploration exist such as entropy regulation \cite{mnih_asynchronous_2016}, diversity-driven exploration \cite{hong_diversity-driven_nodate}, or random network distillation \cite{ burda_exploration_2018}. Nevertheless, none of them  are forcing policies to wander far off from the reward path which can be sometimes a source of deception due to poor reward design.

Another major issue with RL appears when one deals with environments with scarce rewards. In such a context, the trained agent needs to perform a lot of actions between two reward signals. For this reason, it can become challenging for the agent to connect the dots and understand the internal environment mechanisms. It is now common practice to counteract reward scarcity through the addition of \textit{auxiliary tasks} to guide the agent toward the main goal, effectively \textit{reward shaping} \cite{andrew_shaping, jaderberg_reinforcement_2016, jaderberg_human-level_2019}. In general, auxiliary tasks have proven to be efficient, however, they also imply further weight tuning which might be another source of difficulties: how much should the agent prioritize an auxiliary task compared to another? compared to the main goal?

Regarding EC, it has already been verified that it can be a competitive alternative to regular RL techniques by exploiting the reward as a performance objective to optimize. Furthermore, leveraging from a population of agents, EC is much more efficient than RL in terms of exploration. EC can outperform RL in deceptive situations \cite{ such_deep_2018}. Unfortunately, it is known that genetic algorithms tend to suffer from slow convergence when dealing with more complex environments. 

Knowing that diversity entails multiple solution generation, it appears more adapted to maintain a population of agents. This paper briefly reviews several applications of Evolutionary Reinforcement Learning (ERL), a hybrid concept which gathers assets from both RL and EC and derived from Population Based Training (PBT) \cite{jaderberg_population_2017}. Specifically, ERL in this paper constitutes a way to leverage from the best of RL and EC in order to better tackle the diversity dilemma.

This paper is organized as follows. Section \ref{works} presents literature review of recent studies about behavior diversification. Specifically, Evolutionary Multi-Objective Game Intelligence (EMOGI) is presented in detail \cite{shen_generating_2020}. Section \ref{formulations} proposes several optimization formulations, besides the original formulation proposed by the authors of EMOGI. Two experiments and their results are then presented in Section \ref{experiments}. Extensive comparisons as well as an analysis of the results of each formulation are provided in Section \ref{analysis}. Finally, concluding remarks are given in Section \ref{remarks}.

\section{Behavior Exploration and Diversification}\label{works}

\subsection{Conventional Study on Behavior Diversification}

Diversity has been more and more of a subject of interest because of various reasons. As already stated, diversity can be pursued as a strategy to overcome local optimums and that is the first goal of divergent search algorithms \cite{lehman_abandoning_2011, pugh_quality_2016, gravina_surprise_2017, conti_improving_2018, gravina_2019_qualitydt}. Novelty search for instance works by evolving agents so as to maximize the behavioral distance between individuals of its population. Diversity can also be simply desired to enhance the exploration of the behavior space, without always having performance in sight \cite{zheng_wuji_2019}.

The generation of heterogeneous behaviors can further be advantageous in dynamic environments \cite{nojima_mobile_2011}. Having many optimized agents for a given task can be exploited as various alternatives to adapt to the changes in the environment. 

Another motivation for diversity in multi-agent environments is improved training stability. The idea comes from the principle of complementarity, which motivates agents of a same team to learn how to optimize their teamwork by not behaving the exact same way. Thanks to the automated hyperparameter tuning of PBT, the ERL approach reached performance that was previously not possible to obtain without heavy and precise manual tuning \cite{jaderberg_population_2017, jaderberg_human-level_2019}. Maintaining a population also showcases increased stability and performance during training compared to the regular self-play approach (i.e., learning by playing against oneself), thanks to the diversification of teammates and opponents. ERL in this context is able to generate diverse policies and largely outperform humans in tasks of high complexity.

Then, because diversity provides broader behavior exploration, seeking different behaviors can eventually lead to unexpected or hard-to-obtain results. In that sense exploiting this property, variety in solutions turns out to be interesting for surprising content creation \cite{gravina_constrained_2016}. That is also what EMOGI is trying to achieve, in order to obtain stirring and rich game AI design \cite{shen_generating_2020}.




\subsection{EMOGI Framework}\label{EMOGI}

\begin{algorithm}
\caption{The EMOGI algorithm}\label{alg:emogi}
\KwData{$n$: the size of the population}
\KwResult{$P$: the population of candidates.}\tcc{Randomly initialize}
$P \gets \{\{\pi_\theta^1, R_W^1\}, \dots, \{\pi_\theta^n, R_W^n\}\}$ \;
$Q \gets \emptyset$\;
\Repeat{stop criterion is satisfied}{\tcc*[h]{Asynchronized Evolution}
    $p_1, p_2, \dots, p_u \gets \textit{mating}(P)$\;
    $q_1, q_2, \dots, q_v \gets \textit{crossover}(p_1, p_2, \dots, p_u)$\;
    \For{$q \in \{q_1, q_2, \dots, q_v\}$}{
        $q \gets \textit{evalutate}(\textbf{DRL-Train}(\textit{mutate}(q)))$\;
    }
    $Q \gets Q \cup \{q_1, q_2, \dots, q_v\}$\;
    \tcc{synchronized Selection}
    \If{$|Q| \geq n$}{
        \tcc*[h]{Select sparse candidates}
        $P \gets \textbf{Diverse-Select}(P \cup Q, n)$\;
        $Q \gets \emptyset$\;
    }
}
\Return{$P$}\;
\end{algorithm}

Obtaining a heterogeneous behavior-wise population might sound trivial to achieve at first, but in truth, there is a major difference between diversifying the parameters (genotype) and the behavior (phenotype). Genetic algorithms should have no issues exploring over the options for the weights of the policy network $\pi$ by applying the regular crossover, mutation and selection operations. However, such an extensive genotype exploration does not necessarily implicate good phenotype exploration. Because the phenotype is strictly encoded by the genotype through $\pi$, we hardly know how the behavior is impacted when altering the genes.

In RL, agents usually explore an environment through trial and error, while remaining somewhat uncertain in the choices of their actions through entropy regulation, so as to not discard most options prematurely. The genes are being updated, in order to maximize the future expected returns. From those aspects, RL explores the behavior space suffering less of the black box effect of neural networks. Since we directly update the genotype through the reward function, we know that the network progressively grasps what is ``good'' or ``bad''. That being said, properly designing the function reward to successfully guide the agent toward a targeted policy is another challenge. When dealing with non-trivial environments, the reward function has to be crafted and tuned by hand requiring high domain knowledge. Since the agent training is fundamentally connected to its reward function $R$, its design must be a serious consideration.

There comes the idea of exploration through auxiliary tasks weight tuning \cite{jaderberg_human-level_2019, shen_generating_2020}. In a similar way to hyperparameter tuning achieved with regular PBT, we can explore the reward shaping by tuning the importance of the auxiliary tasks. Such a mechanism not only allows agents to automatically tune their reward function without human intervention, it also enables the generation of different strategies depending on the obtained reward shaping. To this effect, the reward function $R$ can be redefined as follows:

\begin{equation}
R= r_{perf} \cdot w_{perf} + \sum_{i=1}^{N} r_i \cdot w_i,\label{reward_shaping}
\end{equation}
where $r_{perf}$ and $w_{perf}$ are the reward value and importance of the main goal which leads to performance, respectively. $N$ is the number of auxiliary tasks, $r=\{r_1,...,r_N\}$ the auxiliary rewards, and $w=\{w_1,...,w_N\}$ their corresponding importance. Finally, $W = w \cup \{w_{perf}\}$ is the obtained reward shaping.

In this manner, the reward can be shaped in many ways, each leading to different behaviors. Taking the example of a boxing game, if an agent puts higher priority in hitting compared to avoiding hits, it  ends up learning to behave aggressively. In the opposite case, the reward shaping has to guide the agent toward a defensive play-style.

EMOGI is an ERL framework, see  Algorithm~\ref{alg:emogi},  adopting the automated reward shaping mechanism coupled with Multi-Objective Optimization (MOO) in order to obtain various desired behaviors in single-agent environments. It first exploits the reward shaping evolution as a way to guide individuals toward targeted strategies through auxiliary rewards. Then, it executes MOO when selecting individuals after each iteration, using behavior objectives that are intrinsically linked to an auxiliary reward. Since an auxiliary task is guiding agents toward a specific behavior, behavior objectives must be chosen so as to reliably evaluate a policy regarding the same behavior.

Getting back to the above boxing example, one could target aggressive and defensive play-styles. Thus, two auxiliary rewards $r_{agg}$ and $r_{def}$, goals that motivate aggressiveness and defensiveness respectively can be chosen. From here, we have to define two behavior objectives $\zeta_{agg}$ and $\zeta_{def}$, that evaluate the resulting policy of an individual regarding the two desired behaviors.

Altogether, individuals that prioritize a goal linked to a play-style should converge toward that play-style, and then be selected through MOO if their learned behavior has a high degree in the corresponding behavior objective. MOO in particular helps conserving diversity in the population over the iterations, as it prevents the population from converging toward one single play-style.

EMOGI, however, does not solely want diversity in strategies, it also looks for performing solutions that maximizes their winning probability in a game. To that extent, it has a similar goal to quality-diversity algorithms \cite{pugh_quality_2016, conti_improving_2018, gravina_2019_qualitydt}, except that the targeted behaviors are known from the start. In order to obtain performing agents, a reward and an objective directly related to performance are added. The performance objective is then prioritized in the MOO so that agents first maximize performance before improving their diversity.

The choice regarding performance prioritization proposed by the EMOGI framework was justified as a mean to produce complex behaviors. This argument needs to be evaluated in order to understand better its impact on the population evolution across the generations. Several alternative formulations are proposed in Section \ref{formulations}, the EMOGI original optimization formulation being given in Section \ref{MOO_EMOGI}.




\section{Formulation of the Optimization Problems}\label{formulations}

\begin{figure*}
\begin{subfigure}{.33\textwidth}
  \centering
  \includegraphics[width=.99\linewidth]{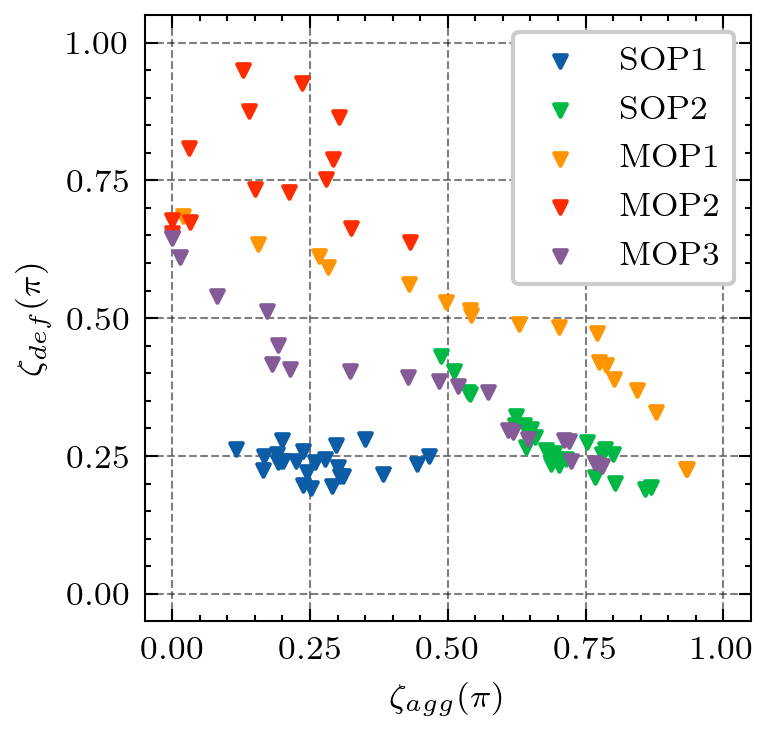}
  \caption{Best run.}
  \label{fig:boxing_behavior_scatter_best}
\end{subfigure}%
\begin{subfigure}{.33\textwidth}
  \centering
  \includegraphics[width=.99\linewidth]{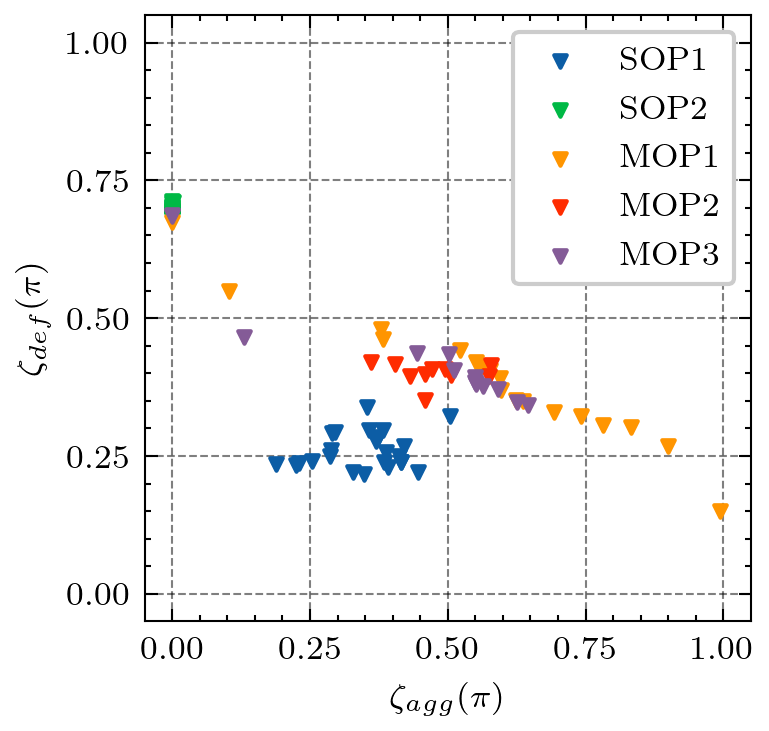}
  \caption{Median run.}
  \label{fig:boxing_behavior_scatter_median}
\end{subfigure}
\begin{subfigure}{.33\textwidth}
  \centering
  \includegraphics[width=.99\linewidth]{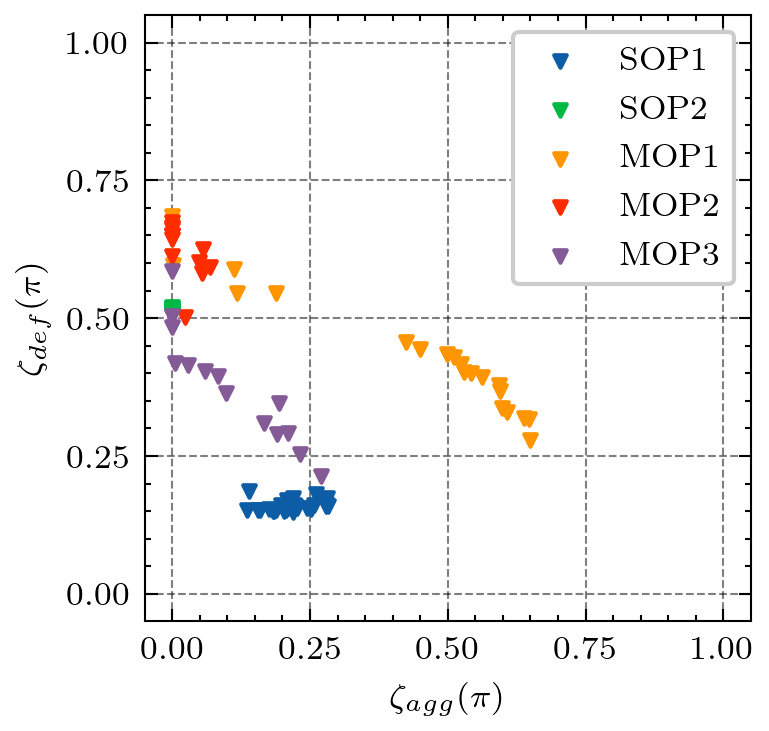}
  \caption{Worst run.}
  \label{fig:plots/boxing/boxing_behavior_scatter_worst}
\end{subfigure}
\caption{Obtained behaviors for Atari Boxing after 50 iterations over six distinct runs. }
\label{fig:boxing_behavior_scatter_all}
\end{figure*}
\begin{table*}[htbp]
\begin{minipage}{\linewidth}
\caption{Average evaluation results, regarding behavior related metrics, for Atari Boxing}
\begin{center}
\begin{tabular}{|c|c|c|c|c|}
\hline
\multirow{2}{*}{\textbf{Problem}} & \multicolumn{3}{c|}{\textbf{Highest objective scores}} & \multicolumn{1}{c|}{\textbf{\textit{Fraction of the population}}}\\
\cline{2-4}
 & \textbf{\textit{Performance}}& \textbf{\textit{Aggressiveness}}& \textbf{\textit{Defensiveness}}& \textbf{\textit{that has a performance of over $0.95$}}\\
\hline
SOP1 & 0.983 & 0.437 & 0.290 & 0.800\\
SOP2 & 0.583 & 0.289 & 0.534 & 0.320\\
MOP1 & \textbf{1.000} & \textbf{0.872} & \textbf{0.726} & 0.813\\
MOP2 & \textbf{1.000} & 0.514 & 0.605 & \textbf{1.000}\\
MOP3 & \textbf{1.000} & 0.638 & 0.633 & 0.507\\
\hline
\end{tabular}
\label{boxing_tab_score}
\end{center}
\end{minipage}
\end{table*}

\subsection{Controls}

To properly observe the impact of more complex formulations, we use two single-objective problems as a reference.

The first single objective formulation Single-Objective Problem 1 (SOP1), focuses on performance and allows better perception of the impact of behavior objectives.

\begin{maxi}|s|
    {\pi}{\zeta_{perf}(\pi)}{}{}.
\end{maxi}\label{eq_SOP1}

The next single-objective formulation, Single-Objective Problem 2 (SOP2), tests the EMOGI framework while maximizing the linear combination of the behavior objectives. This formulation ignores performance, and is used to evaluate the advantage of using MOO.

\begin{maxi}|s|
    {\pi}{\sum_{\zeta \in Z}\zeta(\pi)}{}{},
\end{maxi} 
where $Z = \{\zeta_{{behav}_1},...,\zeta_{{behav}_N}\}$ is the set of chosen behavior objectives.
Naturally, the behavior objective scores are normalized before computing the sum, so that every objective gets considered equally.

\subsection{Focusing Solely on Behavior Diversity}

For MOO formulation, we firstly consider the Multi-Objective Problem 1 (MOP1) which only motivates diversity. To put it simply, MOP1 replaces the sum of SOP2 with the multi-objective dominance operator.

\begin{maxi}|s|
    {\pi}{\zeta_{{behav}_1}(\pi), \dots, \zeta_{{behav}_N}(\pi)}{}{}.
\end{maxi}\label{eq_MOP1}

Compared to the single-objective formulations, MOP1 allows individuals with extreme degrees regarding one behavior objective to be conserved regardless of their score in other objectives. Whence this formulation should preserve heterogeneity better.

\subsection{The EMOGI Approach}\label{MOO_EMOGI}

EMOGI proposes Prioritized Multi-Objective Optimization (PMOO). In PMOO, each objective of the multi-objective formulation are themselves composed of multiple objectives, each ranked by priority. If we consider the complex objective $\zeta(\pi) = [\zeta_1(\pi), \zeta_2(\pi)]$, where $\zeta_1$ is prioritized before $\zeta_2$, the domination relation is changed, so that $\pi_1 \succ \pi_2$ for $\zeta$ if and only if:
\begin{equation}\label{domination1}
   \zeta_1(\pi_1) > \zeta_1(\pi_2),
\end{equation}
or,
\begin{equation}\label{domination2}
    \zeta_1(\pi_1) = \zeta_1(\pi_2),\: \zeta_2(\pi_1) > \zeta_2(\pi_2).
\end{equation}

The justification of EMOGI to use PMOO is to produce complex behaviors, that both learn to follow a particular strategy while maximizing performance. From this point onward in the paper, the EMOGI optimization problem is designated as the second multi-objective formulation, namely, Multi-Objective Problem 2 (MOP2).

\begin{maxi}|s|
    {\pi}{[\zeta_{perf}(\pi), \zeta_{{behav}_1}(\pi) ], \dots, [\zeta_{perf}(\pi), \zeta_{{behav}_N}(\pi) ]}{}{}.
\end{maxi}\label{eq_MOP2}

From the above domination relation, it can already be stated that if the prioritized objective $\zeta_{perf}$ always verifies the first relation defined in \eqref{domination1}, the formulation of MOP2 becomes similar to the one from SOP1.

\subsection{Considering Performance and Diversity Without Prioritization}

The third formulation given by Multi-Objective Problem 3 (MOP3) brings back the concern of MOP2 with performance, but without any form of prioritization.

\begin{maxi}|s|
    {\pi}{\zeta_{perf}(\pi) ,\zeta_{{behav}_1}(\pi), \dots, \zeta_{{behav}_N}(\pi)}{}{}.
\end{maxi}\label{eq_MOP3}

MOP3 is used to evaluate the requirement of the performance objective in the optimization task, as well as understand whether the prioritization hack from EMOGI improves results or not.

\section{Computational Experiments on Some Atari Games} \label{experiments}

\subsection{Atari Boxing}

In order to benchmark the different formulations efficiently, we chose to use the Atari Boxing\footnote{Demonstration video of the obtained individuals in Boxing with MOP1 (our agent in white): \href{https://youtu.be/\_xWT2VYpvXA}{\text{https://youtu.be/\_xWT2VYpvXA}} .} game as a first environment. Boxing lets a player fight against the in-game AI, where the first to punch the other a certain amount of times wins. In the case of timeout (1 minute), we take for winner the player who scored the most points. Each formulation were ran for a total of six times, over 50 iterations. The populations obtained by the  single-objective problems were ranked by their scalar value while the multi-objective problem results were ranked by their hypervolume scores \cite{zitzler_hypervolume_1998}, which are given in Table~\ref{boxing_hv}.

\begin{table}
\begin{minipage}{\linewidth}
\caption{Multi-objective formulation hypervolume scores for Atari Boxing}
\begin{center}
\begin{tabular}{|c|c|c|c|}
\hline
\multirow{2}{*}{\textbf{Problem}} & \multicolumn{3}{c|}{\textbf{Hypervolume scores}} \\
\cline{2-4}
 & \textbf{\textit{Best}}& \textbf{\textit{Median}}& \textbf{\textit{Worst}} \\
\hline
MOP1 & \textbf{0.65} & \textbf{0.52} & \textbf{0.45} \\
MOP2 & 0.56 & 0.39 & 0.14 \\
MOP3 & 0.50 & 0.39 & 0.17 \\
\hline
\end{tabular}
\label{boxing_hv}
\end{center}
\end{minipage}
\end{table}

\begin{figure*}
\begin{subfigure}{.33\textwidth}
  \centering
  \includegraphics[width=.99\linewidth]{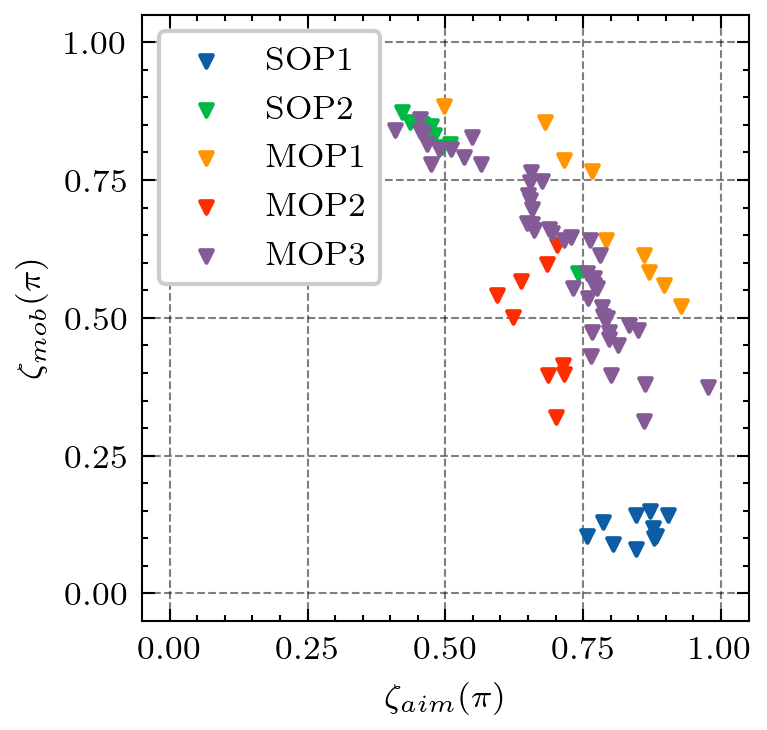}
  \caption{Best run.}
  \label{fig:tennis_behavior_scatter_best}
\end{subfigure}%
\begin{subfigure}{.33\textwidth}
  \centering
  \includegraphics[width=.99\linewidth]{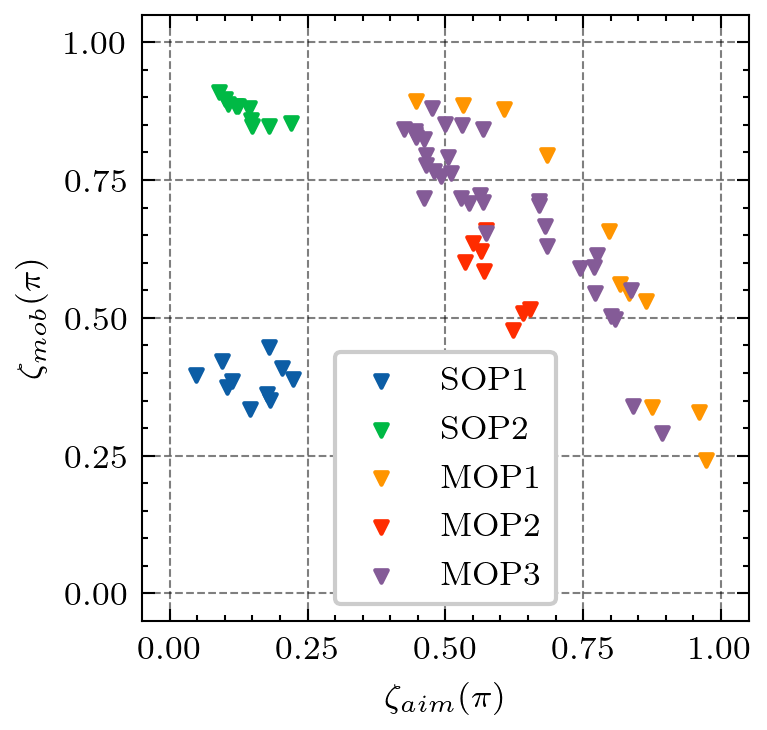}
  \caption{Median run.}
  \label{fig:tennis_behavior_scatter_median}
\end{subfigure}
\begin{subfigure}{.33\textwidth}
  \centering
  \includegraphics[width=.99\linewidth]{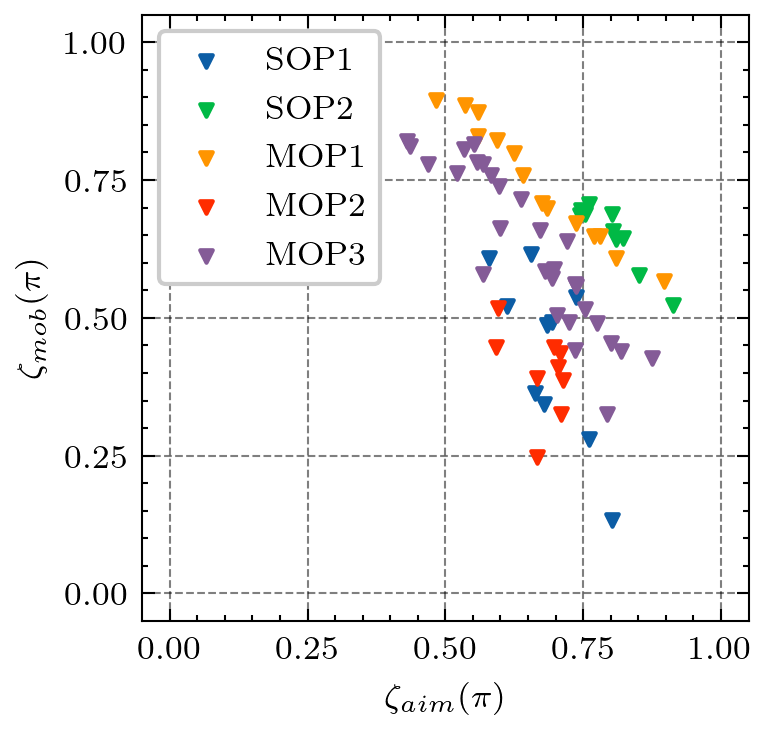}
  \caption{Worst run.}
  \label{fig:tennis_behavior_scatter_worst}
\end{subfigure}
\caption{Obtained behaviors for Atari Tennis after 30 iterations over six distinct runs.}
\label{fig:tennis_behavior_scatter_all}
\end{figure*}
\begin{table*}[htbp]
\begin{minipage}{\linewidth}
\caption{Average evaluation results, regarding behavior related metrics, for Atari Tennis}
\begin{center}
\begin{tabular}{|c|c|c|c|c|}
\hline 
\multirow{2}{*}{\textbf{Problem}} & \multicolumn{3}{c|}{\textbf{Highest objective scores}} & \multicolumn{1}{c|}{\textbf{\textit{Fraction of the population}}}\\
\cline{2-4} 
 & \textbf{\textit{Performance}}& \textbf{\textit{Aim}}& \textbf{\textit{Mobility}}& \textbf{\textit{that has a performance of over $0.5$}}\\
\hline
SOP1 & 0.484 & 0.661 & 0.483 & 0.500\\
SOP2 & 0.651 & 0.631 & 0.863 & 0.234\\
MOP1 & 0.839 & \textbf{0.929} & \textbf{0.898} & 0.242\\
MOP2 & 0.750 & 0.794 & 0.598 & \textbf{0.750}\\
MOP3 & \textbf{0.854} & 0.899 & 0.860 & 0.398\\
\hline
\end{tabular}
\label{tennis_tab_score}
\end{center}
\end{minipage}
\end{table*}

Reproducing the example that was already introduced in Section \ref{EMOGI}, the following auxiliary rewards are used:
\begin{equation}
    r = \{r_{hit}, r_{hurt}\},
\end{equation}
where $r_{hit}$ is the reward motivating the agent to hit its opponent ($r_{hit}=1$ if the player managed to land a punch, $0$ otherwise), $r_{hurt}$ is the reward teaching the agent to avoid punches ($r_{hurt}=-1$ if the player gets hit, $0$ otherwise). Alongside those rewards, $r_{perf}$ is used to motivate performance, it is the reward linked to victory ($r_{perf}=1$ if the player won, $0$ if there is a draw, $-1$ otherwise)

To pair up with those auxiliary tasks, we measure the behavior of our agents with the following behavior objectives:
\begin{equation}
    Z = \{\zeta_{agg}, \zeta_{def}\}.
\end{equation}

 Here, $\zeta_{agg}(\pi)$ and $\zeta_{def}(\pi)$ evaluate the aggressiveness and defensiveness of the agent following policy $\pi$. Naturally, there is no direct way to measure both objectives. We thus have to manually define the objectives, to make them correctly reflect the desired behaviors. Quite obviously, $r_{hit}$ and $r_{hurt}$ are guiding agents to maximize $\zeta_{agg}$ and $\zeta_{def}$, respectively.

Therefore, $\zeta_{agg}$ is decided to measure average game length, and is maximized when the game length is minimized. $\zeta_{def}$ is designed to compute the average distance from the opponent, and is maximized when that mean distance is maximized. In other words, from our human perception, an agent that maximizes $\zeta_{agg}$ tries to finish the game as fast as possible and learns to behave aggressively. Similarly, an agent that maximizes $\zeta_{def}$ tries to stay as far as possible from the opponent and adopts a defensive strategy.

Finally, the performance evaluation which is done through $\zeta_{perf}$, measures the win probability of the player over 10 simulated matches.

\subsection{Atari Tennis}

As a second experiment, we attempted to adapt our framework to the Atari Tennis\footnote{Demonstration video of the obtained individuals with MOP3 on Tennis (our agent in red): \href{https://youtu.be/lsl-gTeizMI}{\text{https://youtu.be/lsl-gTeizMI}} .} environment. The game, compared to the well known and simple Pong game, adds several dimensions to the game, such as a second axis for movement and a third axis for the ball height. A few other mechanisms such as the serve and the consideration of the number of ball bounces adds on the complexity of the game. To successfully conduct our comparison this time, we retrieved the averaged results after 30 iterations of each optimization formulation over six distinct runs. Similarly to Boxing, single-objective problems were ranked by scalar scores, and multi-objective problems were ranked by hypervolume score, which are given in Table~\ref{tennis_hv}.

\begin{table}
\begin{minipage}{\linewidth}
\caption{Multi-objective formulation hypervolume scores for Atari Tennis}
\begin{center}
\begin{tabular}{|c|c|c|c|}
\hline
\multirow{2}{*}{\textbf{Problem}} & \multicolumn{3}{c|}{\textbf{Hypervolume scores}} \\
\cline{2-4} 
 & \textbf{\textit{Best}}& \textbf{\textit{Median}}& \textbf{\textit{Worst}} \\
\hline
MOP1 & \textbf{0.95} & \textbf{0.94} & \textbf{0.92} \\
MOP2 & 0.58 & 0.38 & 0.20 \\
MOP3 & 0.80 & 0.77 & 0.68 \\
\hline
\end{tabular}
\label{tennis_hv}
\end{center}
\end{minipage}
\end{table}

\begin{figure*}
\begin{subfigure}{.5\textwidth}
  \centering
  \includegraphics[width=.66\linewidth]{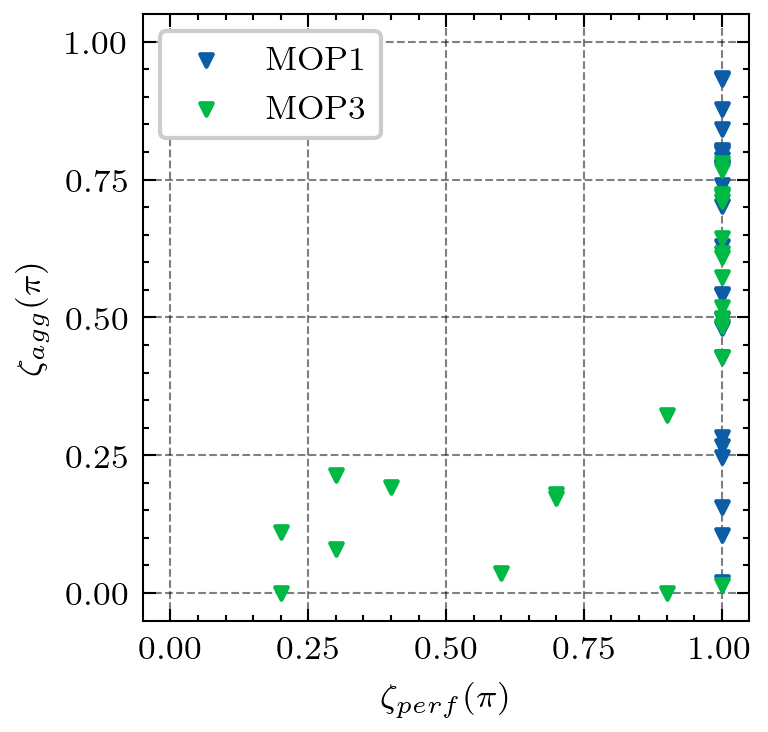}
  \caption{Performance in function of aggressiveness in Boxing.}
  \label{fig:boxing_perf_agg}
\end{subfigure}%
\vspace{11pt}
\begin{subfigure}{.5\textwidth}
  \centering
  \includegraphics[width=.66\linewidth]{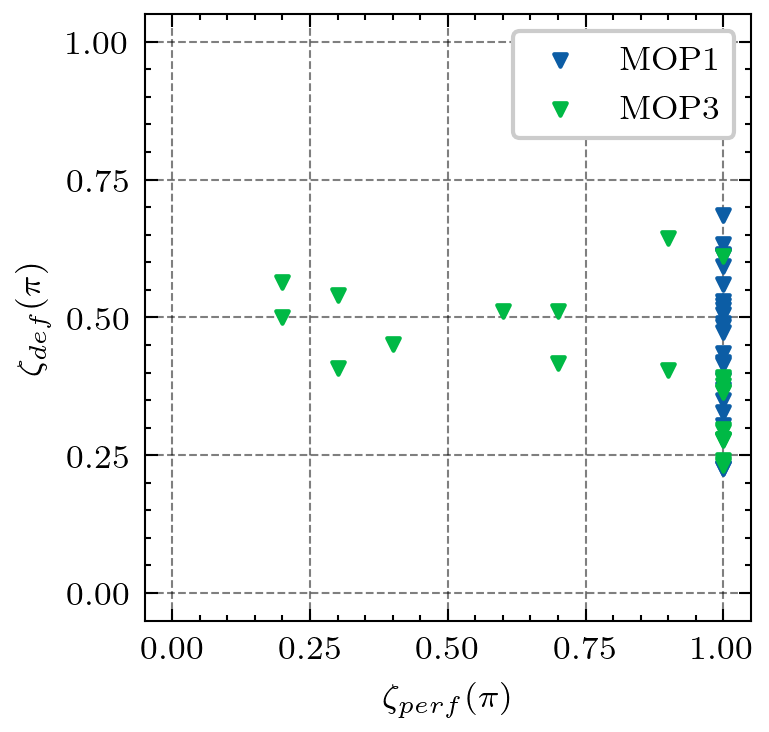}
  \caption{Performance in function of defensiveness in Boxing.}
  \label{fig:boxing_perf_def}
\end{subfigure}
\begin{subfigure}{.5\textwidth}
  \centering
  \includegraphics[width=.66\linewidth]{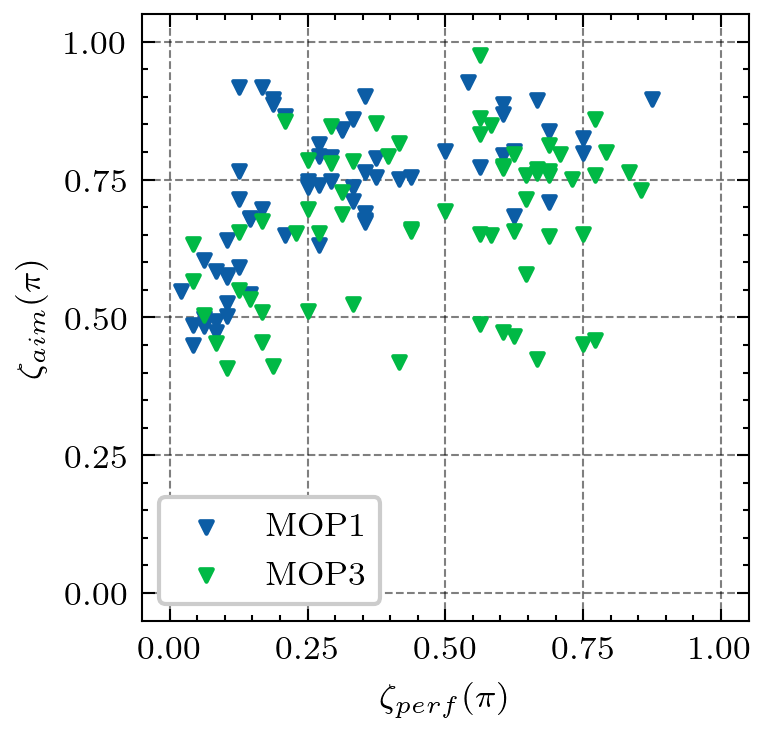}
  \caption{Performance in function of aim quality in Tennis.}
  \label{fig:tennis_perf_aim}
\end{subfigure}
\begin{subfigure}{.5\textwidth}
  \centering
  \includegraphics[width=.66\linewidth]{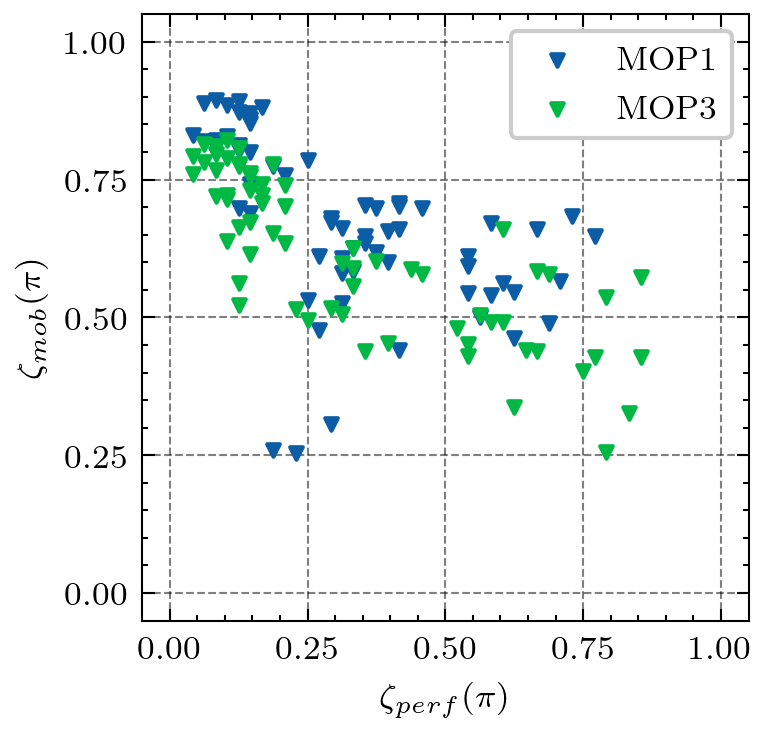}
  \caption{Performance in function of vertical mobility in Tennis.}
  \label{fig:tennis_perf_mob}
\end{subfigure}
\caption{Relations between the performance objective and behavior objectives, in the final population of the best runs for MOP1 and MOP3.}
\label{fig:correlation}
\end{figure*}

For such an environment, we chose to target the following two behaviors: the behavior where the AI tries to hit the ball far from its opponent, and another where the player tries to move as much as possible, between the back and front of the court. In order to achieve such policies, we chose to reward the behaviors with:
\begin{equation}
    r = \{r_{aim}, r_{mob}\},
\end{equation}
where $r_{aim}$ is the reward motivating the agent to maximize the angle of its shots, considering the position of the opponent ($r_{hit}$ being proportional to that angle), and $r_{mob}$ is the reward linked to vertical mobility ($r_{mob}=1$ if the player moves vertically, $0$ otherwise).

Performance is represented by the percentage of rallies won by the agent. Hence, to lead individuals to do the latter, $r_{perf}$ is directly rewarding the agent for winning a rally ($r_{perf}=1$ if the player scored, $-1$ if the opponent scored).

The behavior objectives selecting individuals that successfully managed to learn the targeted behaviors are:
\begin{equation}
    Z = \{\zeta_{aim}, \zeta_{mob}\},
\end{equation}
where $\zeta_{aim}$ and $\zeta_{mob}$ judge the aim quality and vertical mobility, respectively. In order to select individuals with proper aim, $\zeta_{aim}(\pi)$ computes the average angle of shots of policy $\pi$, with some penalty if the AI misses the ball. Then, individuals that are mobile and maximize $\zeta_{mob}$ traveled a lot of distance during the match, thus $\zeta_{mob}(\pi)$ computes the distance traveled per game frame by the agent following policy $\pi$, in order not to penalize individuals that may finish a match faster.

\section{Analysis} \label{analysis}


\subsection{Single-Objective Formulations}\label{sop_result}

According to the obtained results, it is fair to state that single-objective formulations failed to generate any diversity. Figs.~\ref{fig:boxing_behavior_scatter_all} and \ref{fig:tennis_behavior_scatter_all} show that on almost every single run, SOP1 and SOP2 ended up generating a dense behavior cluster, which is definitely not the original goal of the EMOGI algorithm. On top of low behavior diversity, the two formulations obtained lower performance as well. 

It might be understandable for SOP2 as its optimization task ignores performance. But for SOP1 which focuses on solving the task regardless of behaviors, the performance is worse than any multi-objective formulations. This observation is most likely to be justified by the fact that the selection process eliminates any individuals with behaviors that do not necessarily have good performance statistics. This then results in a poorer behavior space exploration, that apparently is required in order to avoid or overcome some local optimums regarding performance. This first observation already stresses the advantage of policy exploration which is brought by diversity.

Focusing on Figs.~\ref{fig:boxing_behavior_scatter_median} and \ref{fig:plots/boxing/boxing_behavior_scatter_worst}, we observe that the population of SOP2 converged toward a very defensive behavior. A similar phenomena is observed in Figs.~\ref{fig:tennis_behavior_scatter_best} and \ref{fig:tennis_behavior_scatter_median} with the mobile behavior. This must be explained by how easy it is to achieve a behavior compared to another. The formulation of SOP2 being single-objective only lets individuals converge toward the simpler behavior. Indeed, defensiveness in Boxing is easy to achieve early on since it only requires agents to run away from their opponent. On the opposite, aggressiveness requires the agent to understand how to play the game in order to beat its opponent before timeout. The same observation can be seen in Tennis since mobility is easier to achieve compared to good aim. Any individual can simply learn how to move back and forth while ignoring the position of its opponent and the ball. An opponent with a good aim though has to consider many more factors and thus is harder to obtain. Based on this statement, SOP2 seemingly tends to converge toward the target behavior that is the most accessible and easy to achieve. 

\subsection{Multi-Objective Formulations}

First of all, Tables~\ref{boxing_tab_score}--\ref{tennis_hv} highlight MOP1 as the best formulation overall. MOP1 has the most consistent performance,  diversity and hypervolume score across experiments. Surprisingly enough, Table~\ref{boxing_tab_score} indicates that MOP1 performs as well as  MOP2 and MOP3 in terms of performance on Boxing, whilst having better diversity even though it does not take in account the performance objective. In Table~\ref{tennis_tab_score}, MOP1 ends up scoring also really high on Tennis, with a slightly lower performance score than MOP3, but much higher scores regarding behavior objectives to compensate.

In Figs.~\ref{fig:boxing_behavior_scatter_all} and \ref{fig:tennis_behavior_scatter_all} looking at the results achieved by MOP2, it is  shown that only a population with lower behavior variety was generated, resulting in clusters looking sometimes like what SOP1 or SOP2 obtained. Overall, MOP2 outputted average diversity on both environments. Tables~\ref{boxing_tab_score} and \ref{tennis_tab_score} show that that the population obtained by MOP2 usually has a better performance across its population. Its maximum performance, however, remains equal or worse than MOP1 and MOP3. Remembering the observation given in Section \ref{MOO_EMOGI}, those results comfort the fact that if equality regarding the performance objective $\zeta_{perf}$ is rare, MOP2 is very close to SOP1 in its functionality.

Focusing the comparison on MOP1 and MOP3, both formulations provide sufficient behavior diversity. What should be stressed however is that, while MOP3 outperforms MOP1 on Tennis regarding overall population performance, it does not on Boxing. Table~\ref{boxing_tab_score} displays a smaller portion of good individuals for MOP3 compared to MOP1, even though MOP1 disregards performance in its formulation. While there can be many other factors justifying this unexpected result, we can mainly blame the behavior objective definition.

In fact, behavior objectives have different impact on the convergence of the population and have different relations with the performance objective $\zeta_{perf}$. For Boxing, Fig.~\ref{fig:boxing_perf_agg} puts forward the fact that high aggressiveness is coupled with high performance. Fig.~\ref{fig:boxing_perf_def} hints that performance is orthogonal to defensiveness. For Tennis, Fig.~\ref{fig:tennis_perf_aim} does also exhibit some positive correlation between performance and aim. But according to Fig.~\ref{fig:tennis_perf_mob}, we observe the opposite between performance and mobility, namely, it gets harder to achieve performance with a more mobile behavior. 

With the case of Boxing with MOP1 performing globally better than MOP3, we can partially explain the result knowing that behavior objectives alone can lead to performance, since we saw they had little to high correlation with performance. Because of that, $\zeta_{agg}$ and $\zeta_{mob}$ are sufficient in order to obtain good agents. MOP3 which adds $\zeta_{perf}$ aside both behavior objectives may limit more the behavior exploration process having larger non-domination frontiers in the later iterations. Such redundancy between objectives may in turns hinder the generation of extreme behaviors with better performance. However on Tennis, the fact that mobility slightly discourages performance explains why MOP3 has more success than MOP1 in terms of performance. The MOP3 population is less likely to be deceived by the mobility factor because it keeps in mind the more reliable performance factor as well.

\section{The Problem with Targeting Specific Behaviors} \label{remarks}

Based on the experimental results in Section~\ref{experiments} and their analysis in Section~\ref{analysis}, we stated that the main issue with the EMOGI framework is that it requires the user to define beforehand the behaviors to be targeted: aggressive, defensive, etc. Regardless of the optimization formulation, having poorly defined behavior objectives and rewards may give uninteresting results. We already saw with Tennis that using behavior objectives that are negatively correlated to performance have an undesired impact on the overall performance. Furthermore, it can eventually be challenging to define and properly adapt some behaviors to the algorithm, depending on the task at hand.

With the EMOGI framework, diversification is performed by trying to dispatch as many individuals as possible across multiple predefined behaviors, but this limits the options for policy diversification. In fact, having two targeted behaviors for instance only provides two convergence directions, effectively generating two extreme strategies as well as intermediary ones. As it is already complex to define behavior objectives, increasing the target number would only increase the difficulty for the framework to generate diversity and performance at the same time \cite{ishibuchi_evolutionary_2008}. 

We could, in a future study, explore the path which no longer targets specific behavior, but aim for a more direct behavior diversity maximization. As it has already been done in the past, by measuring diversity with a behavior-distance metric such as the Kullback-Leibler divergence, a heterogeneous population can be obtained \cite{parker-holder_effective_2020}. Replacing behavior objectives with the diversity metric, we can perform MOO over performance and behavior-distance. On the RL side, many auxiliary tasks could be used, all representing and rewarding a specific event that later helps an individual achieving some unknown behavior.

\section{Conclusion}


This paper proposed four new optimization formulations for the ERL framework of EMOGI. A comparison was established between the original formulation and the ones we proposed. Experiment results successfully stressed the impact of optimization formulation on the behavior diversification process. Single-objective formulations struggle to generate any variety, because they either motivate individuals to converge toward a single optimal point, or fail to consider equally every objectives. The original solution provided by the EMOGI paper was observed to have consistent results in terms of task performance. However, it suffers from the definition of the performance evaluation ending up generating little diversity in solution. In contrast,  the multi-objective formulations that consider objectives without prioritization successfully managed to achieve performance and diversity regarding the targeted behaviors. As such, our proposed multi-objective formulations produced better results both in terms of maximal performance and diversity. Overall, those results successfully confirm the importance of diversity in RL and its benefits over training.

Along our study, we also observed that correlation between achieving a targeted strategies and reach high performance sometimes exist. Because of this possibility, coupling behavior objective optimization with performance optimization is not always the best choice. Furthermore, it can prove to be challenging to properly define behavior objectives and rewards depending on the chosen behaviors. As a future research topic, therefore, pursuing diversity by maximizing some behavioral-distance metric instead of targeting precise behaviors seems to be a good option based on the previous statement.

\bibliographystyle{IEEEtran}
\bibliography{IEEEabrv,references}

\end{document}